\documentclass[runningheads]{llncs}
\usepackage{graphicx}
\usepackage{color}

\begin{document}
\title{UPREVE: An End-to-End Causal Discovery Benchmarking System\thanks{This work is supported through an NSF grant \#2200140 ``PIPP Phase I: Predicting Emergence in Multidisciplinary Pandemic Tipping-points (PREEMPT)''and by the US Army Corps of Engineers ``Engineering With Nature Initiative'' through Cooperative Ecosystem Studies Unit Agreement \#W912HZ-21-2-0040.}}
%
%
\author{Suraj Jyothi Unni
\and
Paras Sheth
\and
Kaize Ding
\and
Huan Liu
\and
K. Sel\c{c}uk Candan
}
\authorrunning{Jyothi Unni et al.}
%
\institute{Arizona State University, Tempe AZ 85281, USA
\email{\{sjyothiu,psheth5,kding9,huanliu,candan\}@asu.edu}}
\maketitle              
\begin{abstract}
Discovering causal relationships in complex socio-behavioral systems is challenging but essential for informed decision-making.
We present {\em \underline{U}pload, \underline{PRE}process, \underline{V}isualize, and \underline{E}valuate} (UPREVE), a user-friendly web-based graphical user interface (GUI) designed to simplify the process of causal discovery. UPREVE allows users to run multiple algorithms simultaneously, visualize causal relationships, and evaluate the accuracy of learned causal graphs. With its accessible interface and customizable features, UPREVE empowers researchers  and practitioners in social computing and behavioral-cultural modeling (among others) to explore and understand causal relationships effectively. Our proposed solution aims to make causal discovery more accessible and user-friendly, enabling users to gain valuable insights for better decision-making.

\end{abstract}
\section{Introduction}
Discovering causal relationships in complex socio-behavioral systems is challenging but essential for informed decision-making:
\begin{itemize}
\item {\em Motivating application $\#1$: Pandemic Preparedness:} Even though they involve biological (such as viruses, vaccines) and physical (such as PPE) components, pandemics are largely socio-behavioral phenomena. 
Therefore,  capturing the spatiotemporal evolution of pandemics at different spatial scales through human interactions, assessing the effects of travel controls during the early epidemic phase, predicting the effects of school closures in mitigating disease spread, assessing the impact of social distancing and  vaccination strategies, and identifying the effects of private responses to evolving disease threats, all require an understanding of the fundamental relationships among social, behavioral, and economic factors impacting human and community behavior, behavior change, in the context of other factors impacting disease spread.

\item {\em Motivating application $\#2$: Water Sustainability:} Even though they involve natural (such as rivers, lakes) and built (such as dams, canals, reservoirs) components, water sustainability is a largely socio-behavioral phenomenon. The amount of water individuals and communities use for domestic, industrial, and agricultural applications can substantially impact water availability. In particular, land use practices such as urban development, deforestation, and agriculture can impact water sustainability through increased demand, pollution, and contamination. Therefore, responding to water sustainability challenges requires understanding the fundamental relationships among social, behavioral, economic factors impacting individual and collective behavior and behavior change in the context of other factors impacting water sustainability.

\end{itemize}
In these and other scenarios, causal relationships can help researchers understand the underlying mechanisms that drive social and behavioral phenomena and help design effective policies and interventions. The improved predictive power through causally-informed reasoning can help forecast the potential outcomes of interventions or policy changes. Through counterfactual reasoning, decision-makers can better assess the impacts of those interventions on the underlying human-centered systems. 

Existing causal discovery toolbox libraries, such as causallearn \cite{causal-learn} and causal discovery toolbox \cite{kalainathan2019causal}, often lack user-friendliness due to their terminal-based interfaces. These libraries may also have limited algorithm and metric options, hindering their practical utility. To address these limitations, we propose  {\em \underline{U}pload, \underline{PRE}process, \underline{V}isualize, and \underline{E}valuate} (UPREVE\footnote{UPREVE is available under MIT License at https://github.com/surajjunni/UPREVE.}), a graphical user interface (GUI) for causal discovery. UPREVE aims to provide a user-friendly environment for algorithm selection, data visualization, and result evaluation by incorporating various state-of-the-art algorithms and metrics. In comparison to CastleBoard \cite{zhang2021gcastle}, an existing GUI, UPREVE offers the advantage of running multiple algorithms simultaneously on a single dataset, along with data preprocessing capabilities. UPREVE also provides a wider range of visualization options, including heatmaps and directed graphs, allowing users to gain deeper insights into causal relationships. Additionally, UPREVE supports multiple users, fostering collaboration and concurrent usage. By introducing UPREVE, we aim to overcome the limitations of existing libraries and offer a comprehensive solution for efficient and insightful causal discovery. The subsequent sections will delve into UPREVE's modules, present a case study using the Covertype dataset, analyze the results, discuss limitations, and outline potential avenues for future development.

\begin{figure}
    \centering
    \includegraphics[scale=0.55]{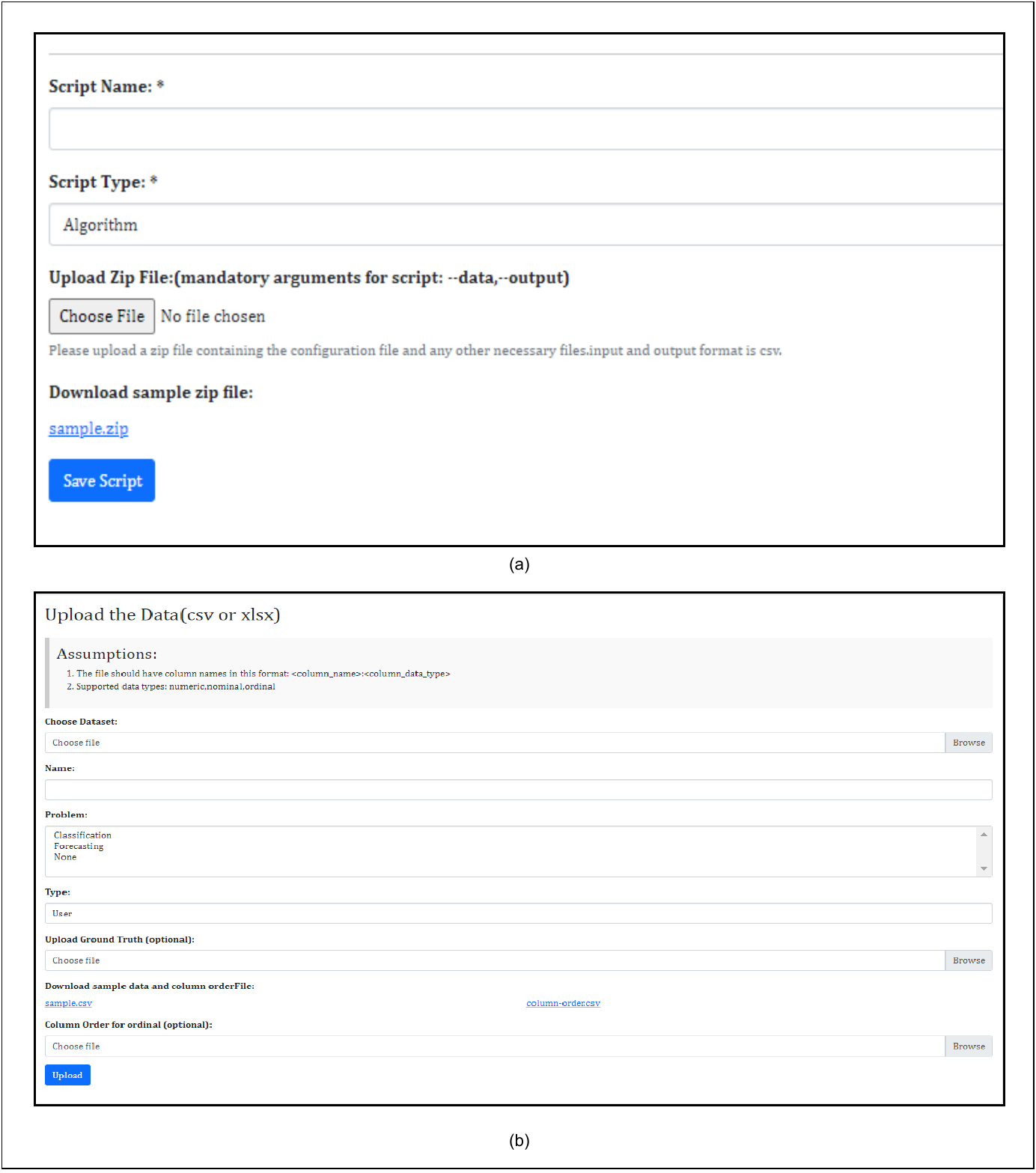}
    \caption{(a) Algorithm/Metric Upload Module,(b) Dataset Upload and Preprocess Module}
\end{figure}
\begin{figure}
    \centering
    \includegraphics[scale=0.5]{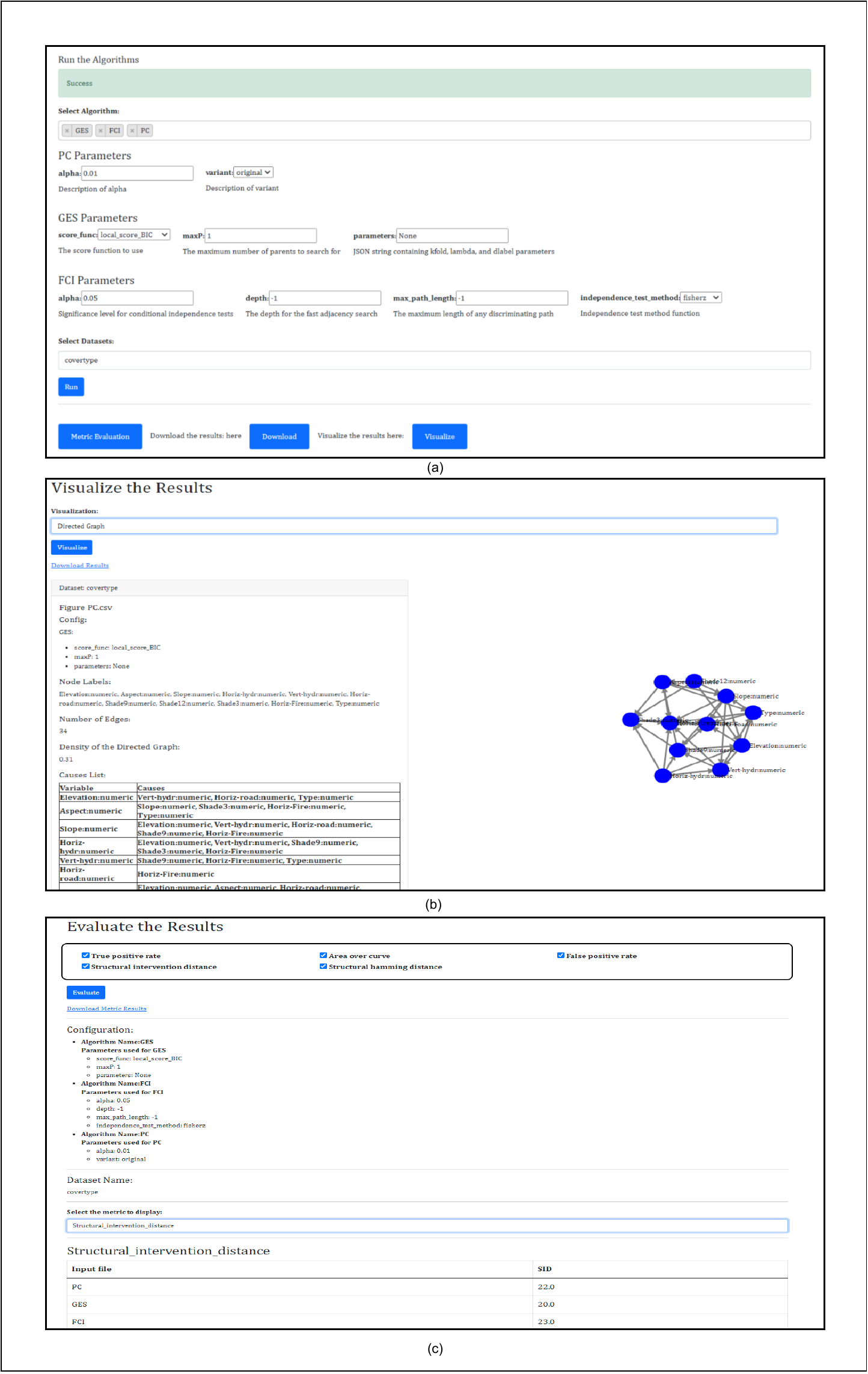}
    \caption{(a) Algorithm Execution Module, (b) Metric Evaluation Module, (c) Algorithm/Metric Upload Module}
  
\end{figure}
\section{UPREVE Architecture}
The UPREVE system facilitates the causal discovery and analysis of complex datasets. It consists of several modules that interact with each other to streamline the process of causal discovery. The following modules are integral to the UPREVE architecture:

\begin{itemize}
    \item \textbf{Dataset Upload and Preprocess Module} allows users to upload CSV/XLSX datasets into the UPREVE system. The module prompts the user to provide information about the dataset described in the case study section. Additionally, the module requires the user to provide ground truth data calculated by domain experts in CSV format. This module supports datasets with column names in the format \textit{$<$columnname$>$:$<$column type $>$}, enabling proper identification of variables.
    \item \textbf{Algorithm/Metric Upload Module} enables users to upload scripts related to causal algorithms and metric evaluation into the UPREVE system. Users can upload pre-existing algorithms or create their own by uploading a zip file. The zip file should include a \textbf{config.json} file specifying the algorithm's name, parameters, script language, and script path and a \textbf{requirements.txt} file listing any necessary package dependencies. The module supports Python and R scripts for algorithm execution. Similarly, metrics evaluation scripts can also be uploaded in the same way.

    \item \textbf{Algorithm Execution Module} executes causal discovery algorithms on the uploaded dataset. It supports several cutting-edge causal discovery algorithms([1],[2],[3],[5],[6],[7]) within the system.Users can select the algorithms they wish to execute and adjust their configurations suitably. Additionally, users can choose the dataset on which the algorithms will be applied. Upon submitting the selection, the module generates results that can be downloaded, visualized, or evaluated using appropriate metrics if a ground truth is available. The console log view on the same page displays all relevant logs and error messages related to the execution process.
\textbf{Visualization Module} enables users to observe and interact with the outcomes of the executed algorithms. Users have various graphical options, including directed graphs and heat maps to visualize the inferred causal graph. The module provides metadata for each visualization, such as node names, the number of edges, and the graph's density.Users can download the visualized data for further analysis.

    \item \textbf{Metrics Evaluation Module} allows users to compare the performance of different algorithms for determining causal relationships. The module supports both pre-existing and custom metrics. Currently, it supports five benchmark metrics: area over curve, true positive rate, structural hamming distance, structural intervention distance, and false positive rate. The module presents the evaluation findings in a tabular format, allowing users to select the desired metric and view the corresponding results.
\end{itemize}

Integrating these modules, the UPREVE system provides a comprehensive approach to causal discovery. It automates algorithm selection, provides visualization capabilities for exploring causal relationships, and offers metric evaluation to compare algorithm performance. This architecture empowers researchers and practitioners from various fields to efficiently analyze complex datasets and gain valuable insights for decision-making.
\section{Case Study}
This case study analyzes a forest cover-type dataset using the UPREVE system. The dataset comprises 581,012 instances and consists of 54 columns of data, including quantitative variables, such as elevation, aspect, slope, distances to hydrology, roadways, fire points, and binary wilderness area and soil type designations.
To apply the UPREVE system to the forest cover type case study, we followed a systematic approach that involved several modules.
\subsection{Dataset Upload and Preprocess Module}
The first step is to upload the dataset to the UPREVE system. The dataset is in CSV format. The UPREVE system automatically detects the column names and data types. The module asks for information about the dataset, such as its name, type, and problem. The dataset name for this data is \textbf{covertype}. The type can be \textbf{user} or \textbf{benchmark}. The problem can be \textbf{classification}, \textbf{forecasting}, or \textbf{none}. The problem types list will be increased in the future. The module then asks for ground truth, calculated by domain experts and in CSV format. There is one more option column called "order" for ordinal, but we do not need it because there is no ordinal column in this dataset.

\subsection{Algorithm/Metric Upload Module}
For the Algorithm/Metric Upload Module, we assume that all the required algorithms and metrics for the case study have already been uploaded into the system. These algorithms and metrics follow the same upload procedure as outlined in Section 2.

\subsection{Algorithm Execution Module}
Next we move on to the execution phase, where we experiment with six algorithms (PC\cite{spirtes2000causation}, FCI \cite{DBLP:journals/corr/abs-1302-4983}, VARLiNGAM \cite{6627}, GIN \cite{DBLP:journals/corr/abs-2010-04917}, GES \cite{10.1162/153244303321897717}, and ICALiNGAM \cite{10.5555/1248547.1248619}) with predefined configurations. These algorithms are uploaded in the same way as mentioned in Section 2. In the "home" section, we have the option of selecting these algorithms with their configuration, and after that, we have the option of selecting the dataset, which in this case will be covertype. After that, all six algorithms will be run on the covertype dataset. If there is an error when running, it will be visible in the console log, which is available below this module. Once completed, the system displays three options: visualize, metric evaluation, and download the result. The first two will be covered in later sections.

\subsection{Visualization Module} The Visualization Module in the UPREVE system enables the visualization of the results obtained from the causal discovery algorithms. Two visualization options are available: a directed graph representing the causal relationships between the variables and a heat map illustrating the correlation among the variables in the Forest Cover Type dataset.
Users can download the visualization data for further analysis and exploration. For example, when visualizing the directed graph generated by applying the PC algorithm to the Forest Cover Type dataset, the visualization displays node names, the number of edges (34), the graph's density (0.31), and a table that lists the causes of each node as shown in Figure 2b. In the case of the node \textbf{\textit{Elevation:numeric}}, the contributing nodes include \textbf{\textit{Horiz-hydr}:numeric, \textit{Shade9}:numeric, \textit{Shade12}:numeric, \textit{and} \textit{Shade3}:numeric}. Similar visualization data and causal relationships can be obtained for the other five algorithms in the UPREVE system.
\subsection{Metrics Evaluation Module}
This module efficiently calculates essential metrics for each algorithm, such as the area under the curve, true positive rate, structural Hamming distance, structural intervention distance, and false positive rate. Subsequently, a comprehensive and detailed report is generated, summarizing the outcomes of the metric evaluation.
In the case of the PC algorithm and its structural intervention distance, the evaluation reveals a value of 22 (illustrated in Figure 2c). This signifies that the current configuration employed for the PC algorithm is suboptimal for the given dataset. Consequently, it strongly suggests the necessity of reconfiguring the algorithm or exploring alternative algorithms to enhance performance.
\section{Conclusion and Future Directions}
This paper introduces UPREVE, an interactive graphical user interface (GUI) designed to streamline the causal discovery process for socio-behavioral research and decision making. UPREVE offers a unified system that allows users to execute various causal discovery algorithms on datasets compatible with R and Python. The GUI also provides visualization capabilities to explore the interactions between variables based on algorithm results, facilitating a better understanding of causal relationships. Additionally, UPREVE can identify similarities among algorithm outcomes.
A session-based feature will be incorporated into UPREVE in future development to enable users to track their previous results conveniently. Furthermore, the aim is to explore and experiment with additional functionalities that can provide users with more insights related to the results of causal discovery algorithms. These enhancements will enhance the usability and effectiveness of UPREVE, empowering users in their causal discovery endeavors.

\bibliographystyle{splncs04}
\bibliography{upreve}

\end{document}